\documentclass[letterpaper, 10 pt, conference]{ieeeconf}  

\IEEEoverridecommandlockouts

\usepackage{graphicx}
\usepackage{textgreek}
\usepackage{amsmath}
\usepackage{caption}
\usepackage{subcaption}
\usepackage{booktabs}
\usepackage{multirow}
\usepackage{array}
\usepackage{comment}
\usepackage{longtable}
\usepackage{url}
\usepackage{hyperref}
\usepackage{cite}

\title{\LARGE \bf
UltraBot: Autonomous Mobile Robot for Indoor UV-C Disinfection with Non-trivial Shape of Disinfection Zone
}
\author{\authorblockN{Nikita Mikhailovskiy, Alexander Sedunin, Stepan Perminov, Ivan Kalinov, and Dzmitry Tsetserukou}
\authorblockA{ \textit{Skolkovo Institute of Science and Technology}
\textit{Moscow, Russia 121205}\\
\{nikita.mikhailovskiy, alexander.sedunin, stepan.perminov, i.kalinov, d.tsetserukou\}@skoltech.ru}}

\begin{document}

\IEEEoverridecommandlockouts
\pubid{\makebox[\columnwidth]{978-1-7281-2989-1/21/\$31.00 \copyright2021 IEEE\hfill} \hspace{\columnsep}\makebox[\columnwidth]{ }}

\maketitle

\begin{abstract}
The paper focuses on the development of an autonomous disinfection robot UltraBot to reduce COVID-19 transmission along with other harmful bacteria and viruses. The motivation behind the research is to develop such a robot that is capable of performing disinfection tasks without the use of harmful sprays and chemicals that can leave residues and require airing the room afterward for a long time.
UltraBot technology has the potential to offer the most optimal autonomous disinfection performance along with taking care of people, keeping them from getting under the UV-C radiation. The paper highlights UltraBot's mechanical and electrical design as well as disinfection performance. 
The conducted experiments demonstrate the effectiveness of robot disinfection ability and actual disinfection area per each side with UV-C lamp array. The disinfection effectiveness results show actual performance for the multi-pass technique that provides 1-log reduction with combined direct UV-C exposure and ozone-based air purification after two robot passes at a speed of 0.14 m/s. This technique has the same performance as ten minutes static disinfection. Finally, we have calculated the non-trivial form of the robot disinfection zone by two consecutive experiment to produce optimal path planning and to provide full disinfection in selected areas.

\end{abstract}

\section{Introduction}


The COVID-19 pandemic imparted us a lesson, that the threat is not a one-time phenomenon and that now it is always worth worrying about disinfection and personal protective equipment. The biggest risk zones are that  populated with large crowds of people, i.e., shopping centers, shops, hospitals, and etc. Studies show that COVID-19 stays on various surfaces for up to 80 hours and touchable surfaces \cite{Doremalen2020}, e.g., door handles, handrails, office furniture, and etc., represent a high risk of transmitting the virus. Another problem in enclosed spaces is that COVID-19 is airborne and highly transmittable \cite{Asadi2020}, and it remains in the air in the form of particles for up to 3 hours \cite{Doremalen2020}.

The main problem in the struggle for disinfection is that in most places, at the moment, stationary methods are used, or manual methods of cleaning and disinfecting the surface. Even though manual disinfection is quite effective, since many inaccessible places are cleaned, these methods are also dangerous for the operators. If they are not indoor disinfection devices, stationary methods also require operators, which increases the risk of their use, as they have to be constantly moved according to the disinfection schedule.

One of the most effective disinfection methods today is application of ultraviolet-C light. Research shows that it is effective against SARS-CoV-2 and numerous other harmful bacteria and viruses \cite{Bianco2020, YANG2019487}. Also, ultraviolet light can purify the air within the range of the emitter's disinfectant power.  

\begin{figure}
\centering
\includegraphics[width=0.42\textwidth]{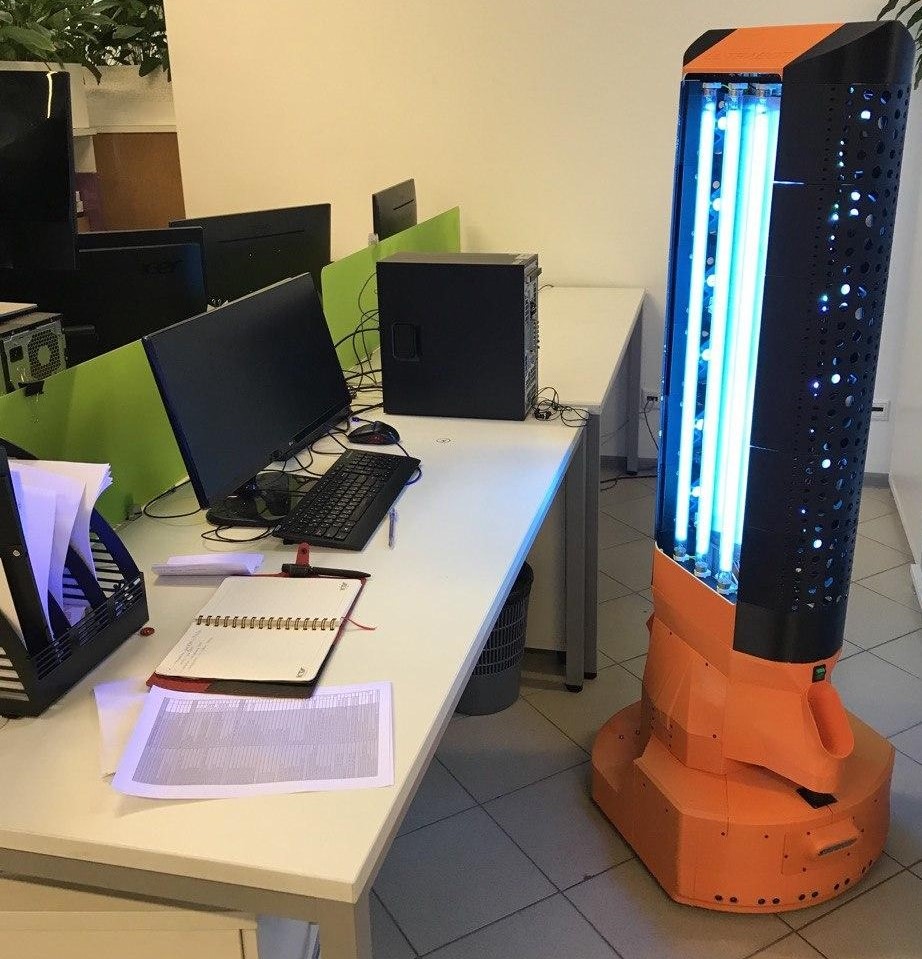}
\caption{UltraBot disinfecting company office.}
\vspace{-1.5em}
\label{UB}
\end{figure}

\pubidadjcol

\subsection{Motivation and problem statement}
A variety of methods to disinfect the premises are known to this day. There are three main ways to treat the surfaces, such as pulverization or vaporization of some liquids, ultraviolet radiation, or physical treatment. All of them have their pros and cons, yet, in search of a universal solution, the ultraviolet-based method of disinfection has its predominant features. Three major parameters that differ the approaches for disinfection are the homogeneous distribution of the substance or radiation, the longevity of the disinfection cycle, and safety to humans. Otter et al. compared unmanned disinfection systems and provided a detailed overview of the virtues and shortcomings of the approaches \cite{OTTER2020323}. In accordance with the research results, the most effective were ultraviolet-based disinfection systems.

Whereas vaporization systems are highly effective for bacteria sanification, they are both dangerous for the human respiratory system. The main consequence of this is that the room needs to be ventilated for a significant time before entry. The second major shortcoming of all liquid disinfectants is their low efficiency against the viruses spread along the surfaces \cite{Kampf2020246}. The third problem, is that such systems are more viable for long use in closed rooms (cycle $\geq$1 hour) and ineffective in huge spaces. Additionally, high concentrations of some compounds in vapor can damage the metal structures. The final issue associated with the systems that use liquids to disinfect the premises is to control the liquids' quality and quantity for system operation and utilization.

There are two types of radiation wavelengths among ultraviolet treatment, known as Ultraviolet-C (UV-C) and Pulse-Xenon Ultraviolet (PX-UV). The research \cite{stibich2011} allows us to assume that PX-UV is an effective way for touchless cleaning of the surfaces and can be applied for premises disinfection. However, Nerandzic et al. presented the comparison of UV-C and PX-UV effectiveness in their research \cite{Nerandzic2015}, according to which UV-C had shown better results, besides some limitations, e.g. lamp disposal. The best way to utilize PX-UV lamps is to co-use manual cleaning with short brakes  \cite{LI2020111869}. However, it makes it impossible to build a completely autonomous system based on PX-UV disinfection.

Current technical solutions for UV-C disinfection of air and surfaces can be divided into mobile and non-mobile. Non-mobile disinfecting systems can be represented by stationary installations with UV-C  lamps, UV cleaner re-circulation systems, and others. A stationary germicidal box with UV-C lamps can clean only air in a small area near the device. However, the research says that while sneezing, large droplets (with diameter of 60 \textmu m - 100 \textmu m) are carried away for more than 6 m of horizontal distance \cite{TransmissionCOVID}. These droplets are too heavy to remain in the air and fall on the floor or nearby surfaces. Consequently, a large space cannot be sanitized without surface treatment by a stationary system. 

Autonomous robotic systems have been gaining popularity in numerous fields of industry, and various tasks from stocktaking in warehouses \cite{kalinov2019high, kalinov2020warevision, kalinov2021Impedance} and shopping rooms \cite{decathlon}, plant disease detection \cite{karpyshev2021autonomous} to cleaning and sanitizing large areas.
Different types of robots can clean the floor and clear the recycle bins, disinfecting public spaces \cite{floor_can}. Autonomy gives two key advantages. First, robots can use more powerful disinfectants that pose a danger to humans. Secondly, their effectiveness does not decrease as the area that they need to sanitize increases. Such robots can potentially substitute humans, who are working within conditions that represent a threat to their health.

The main problem is the lack of mobile robots that would take into account all these factors and have the ability to work with people in the same room without harming them.


\subsection{Related works}


Nowadays, there are already industrial versions of service robots on the market capable of disinfecting different types of places. 
It is possible to divide these robots into several groups based on the area to be disinfected. 

For the treatment of large outdoor areas, ground or air robots equipped with chemical sprinklers are used. Drones from XAG \cite{XAG} and DJI \cite{DJI} companies, mainly used for agricultural purposes, have shown promising results in disinfecting city streets in China \cite{sinha}. 
Despite good outdoor disinfection results, this method is less effective and kills fewer bacteria \cite{Kampf2020246}. Such robots are semi-autonomous, and for each disinfection, they need to be manually set up to a route. Also, these robots need constant replenishment of the disinfectant spray agent. In contrast, the robot developed by us is completely autonomous, and its operating time is limited only by the battery charge.

For indoor disinfection, in compact premises with a high density of people, robots equipped with UV-C lamps are the most effective. This type of robot can be implemented entirely autonomously. Chanprakon et al. presented the first attempt in this direction by basic design of a simple UV-C disinfection robot and study of UV-C irradiation on S. aureus bacteria \cite{RobotUVCIEEE}. Tiseni et al. \cite{tiseni2021uv} as well as main market players UVD Robotics \cite{UVDrobots}  and Honeywell \cite{Honeywell} use a completely open lamp arrangement, which limits their use in only deserted rooms and obliges to control the entire disinfection process. At the same time, UltraBot due to its design and software, does not irradiate directly on people, tracking their positions during its work, and turns off the appropriate set of lamps. Nevertheless, McGinn et al. \cite{mcginn2020exploring},  \cite{tmirob} presented robots improving current design trends and providing disinfection with limited irradiation areas for additional safety and purifying a bounded space.

However, none of the companies covered protocols for a safe scenario of work in a case when people are detected on the way.


\subsection{Contribution}

 In this work, a novel robot UltraBot for autonomous UV-C disinfection considering the room volume and the presence of people nearby is presented and tested for disinfection efficiency. The prototype can disinfect populated premises, e.g., shops, warehouses, and workplaces in multiple passes using half and whole UV-C lamp irradiation zones and applying the safety algorithms. It may track nearby areas for people and instantly disable lamps in the case of their occurrence.  Also, the robot uses additional breezing of ambient air to distribute ozone generated by the lamps, which increases cleaning efficiency and allows disinfecting hard-to-reach areas \cite{microorganisms9010172}.

 {\bf Our main contributions are as follows:}

\begin{itemize}
\item Fully autonomous and safe for people robot for UV-C disinfection of warehouses, shopping malls, open office spaces, campuses, hospitals, and etc. 
\item Various scenarios of disinfection and experimental results.
\item Methodology for constructing a disinfection spot, reflecting the most effective areas to consider for the robot path planning.
\end{itemize}
 

\section{Robot Overview}

\begin{figure}[!t]
\centering

\includegraphics[scale=0.5, width=0.4\textwidth]{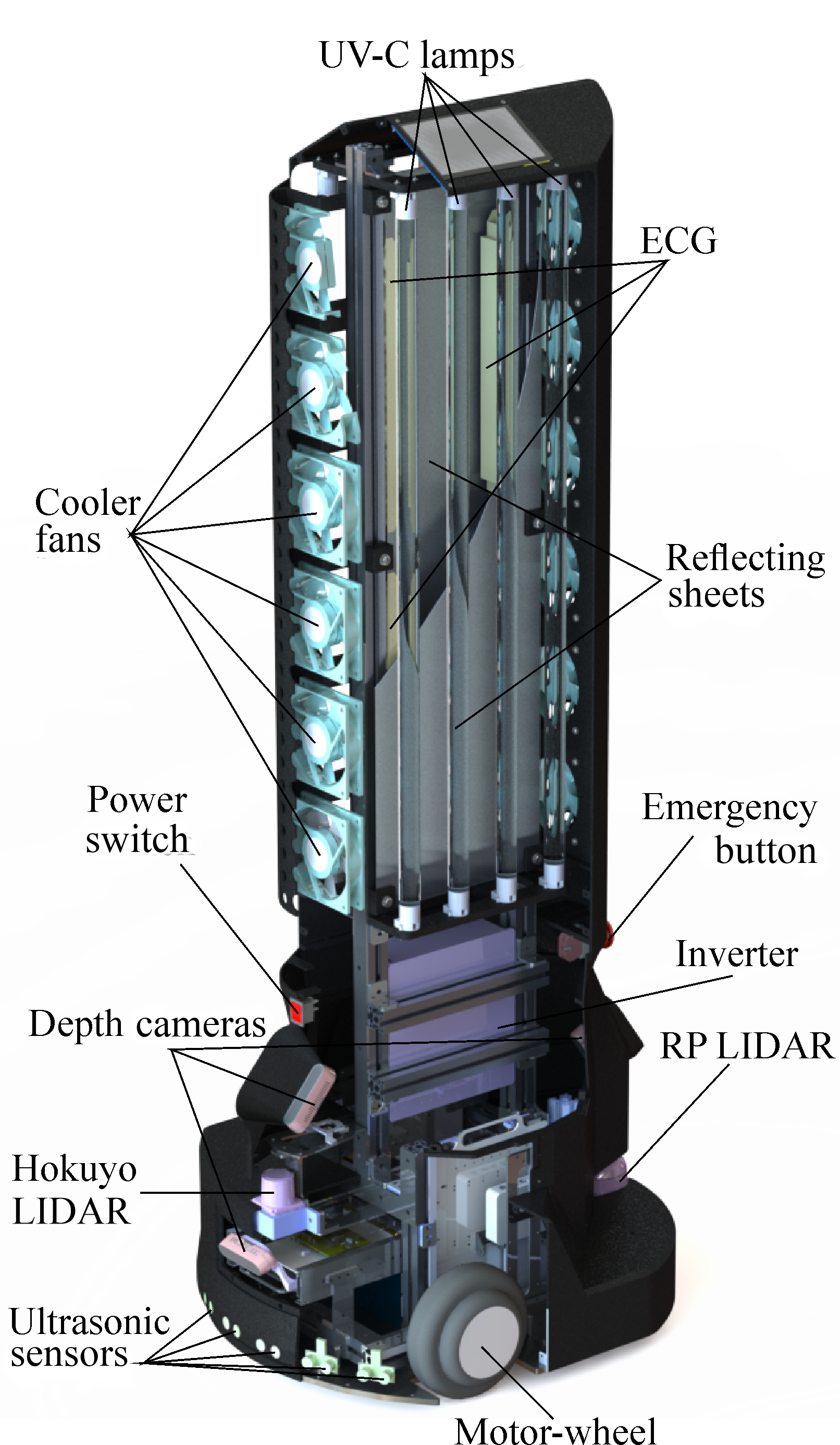}\label{true}
\caption{Decomposition of the robot structure.}
\label{fig:isometric}
\vspace{-2em}
\end{figure}

As a light source for disinfection, UV-C mercury vapor lamps of Phillips TUV series  were used. The UV-C fluorescent lamps are working at 253.7 nm wavelength, which has proven the highest germicidal effectiveness \cite{UVResearch}-\cite{UVCGerm}. Eight lamps are installed on two opposite sides (4 lamps in the stack) of the robot, and each stack can be used separately. Each illuminant has 100 \textmu{W/cm$^2$} of UV-C irradiation at a distance of 1 meter and 30 W of electrical power. Therefore, each side has 400 \textmu{W/cm$^2$} of irradiation and 120 W of power consumption, which is enough to kill 90\% of microorganisms \cite{Philips}. The control is carried out using relays that commute AC voltage to  Electronic Control Gear (ECG) drivers.

We placed a strong attention on the safety of neighboring humans.  In contrast to typical UV-C lamp placement in a cylindrical shape, we split the lamps into two opposite sides of the robot, thus, the UV-C emitting area would be restricted to only 180$^\circ$. Overall, it allows the robot to operate concurrently side by side with humans without downtime.  As a result of the lamp stack placement facing each other, the lamps evidently will not work at full power because of the angle of glowing. The lamp module is supplied with a reflective shield made of celled anodized aluminum to avoid the loss of disinfection power \cite{ultrabot}.

In order to disinfect the air and add the ``safe mode" of work, i.e., when the light-blocking shield fully covers the UV-C bulbs in the future version, the robot was supplemented by an array of 12 DC fans. The fans are located lengthwise of the lamps, providing the significant airflow volume to pass through the maximum possible illuminating area. They are intended to work as the UV sterilizer in the situations when the robot is passing the humans. Thus, there will not be a loss in productiveness. During this work mode, the robot blows air stream through the disinfection system, functioning alike with a quartz lamp, simply maintaining airflow with the ionized particles and disinfecting the air. 

Sensors have been chosen to provide high quality object detection and large sensing area. The Hokuyo LIDAR aims implementation of simultaneous localization and mapping (SLAM) because of high precision and update frequency. It is located in front of the robot's body. The RP2 LIDAR determines rear collision to achieve the robot's 360$^\circ$ field of view (FoV). Ten ultrasonic sensors and four Intel RealSense RGB-D cameras are used for collision detection and obstacle avoidance. Ultrasonic sensors are located beneath LIDARs to detect tiny obstacles such as feet, stairs, and transparent objects. RGB-D cameras are applied to detect the person approaching the upper part of the robot. Human detection is necessary for the emergency shutdown of lamps if someone is near the working robot and for the robot path planning while being safe for human.

The control system is based on the high-level controller (Intel NUC computer) and low-level controller board (STM32). The low-level controller board provides interfaces for ultrasonic sensors, UV-C lamp control, LED control, and battery statistics. 
The communication protocol of high and low-level controllers is implemented with a USB virtual COM interface.

\section{Experimental Data}

In order to validate the performance of the mobile robot in terms of disinfection by UV-C lamps and to choose an optimal technique of disinfection process, three studies have been conducted. 

\subsection{Multi-pass disinfection approach performance}

We propose the multi-pass approach to increase the speed of the disinfection procedure and optimize energy consumption. In this type of disinfection, a robot has been passed several times the same place of disinfection, going around the premises. 

\begin{figure}[!t]

    \begin{subfigure}{0.45\textwidth}
        \includegraphics[width=\textwidth]{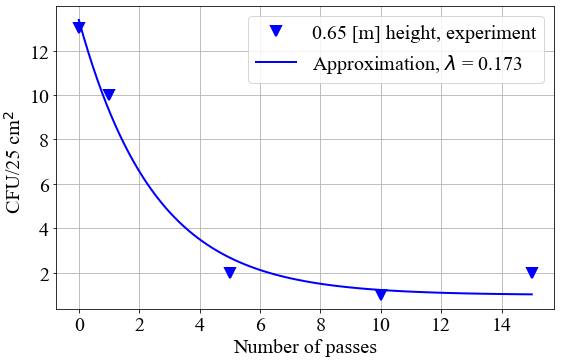}
        \caption{0.35 m distance to the disinfected object}
        \label{tbc_terminal}
    \end{subfigure}
    \begin{subfigure}{0.45\textwidth}
        \includegraphics[width=\textwidth]{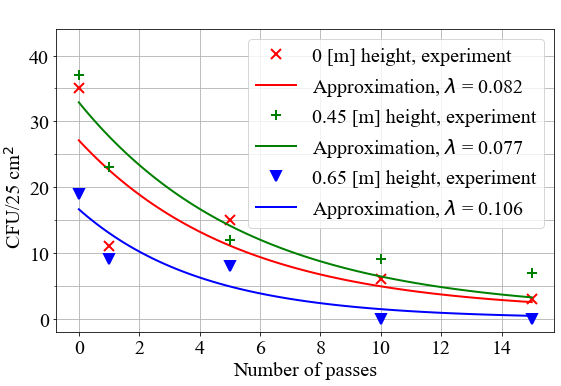}
        \caption{1 m distance to the disinfected objects}
        \label{tbc_surfaces}
    \end{subfigure}
\caption{Results for multi-pass disinfection in souvenir shop.}
\vspace{-1.5 em}
\label{museum}
\end{figure}

The first test series were made in the Russian State Historical Museum at a souvenir shop with turned-off fans to test a direct, far UV-C irradiation effectiveness. 
The robot passed the testing area several times with a constant speed of 0.2 m/s. A specialist took swab samples (in total 15 samples were collected) after each pass from three types of surfaces located at 1 m distance on different heights: floor (0 m), chair (0.45 m), and table (0.65 m). A payment terminal located on the table was additionally tested at 0.35 m distance. The authorized laboratory OLPHARM\footnote{https://olpharm.ru/} has researched the Total Bacteria Count (TBC) of samples providing results in colony-forming units (CFU) per 25 cm$^2$. Experimental results of this research are presented in Fig. \ref{museum}.

To analyze the experimental data, we use the Chick-Watson inactivation kinetics model \cite{peleg2020microbial}: 
\begin{equation}
 N(n) = N_0 10^{-\lambda n}, 
\end{equation}
where $n$ is the number of passes, $N_0 = N(0)$ is the initial CFU/cm$^2$, and the microbial inactivation rate constant $\lambda$ represents a decrease rate of these values due to disinfection. 
This approximation method allows estimating the decrease rate of a total bacterial count numerically and fitting the data optimally.

\begin{figure}[ht]
    \begin{subfigure}{0.45\textwidth}
        \includegraphics[width=\textwidth]{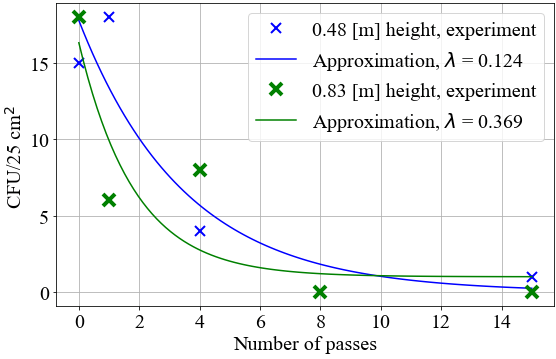}
        \caption{0.3 m distance to the disinfected shelves}
        \label{tbc_03m}
    \end{subfigure}
    
    \begin{subfigure}{0.45\textwidth}
        \includegraphics[width=\textwidth]{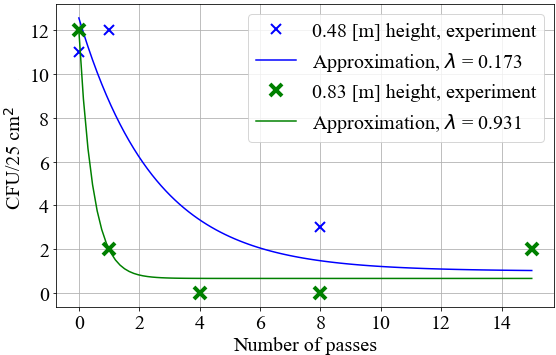}
        \caption{0.6 m distance to the disinfected shelves}
        \label{tbc_06m}
    \end{subfigure}
    \begin{subfigure}{0.45\textwidth}
        \includegraphics[width=\textwidth]{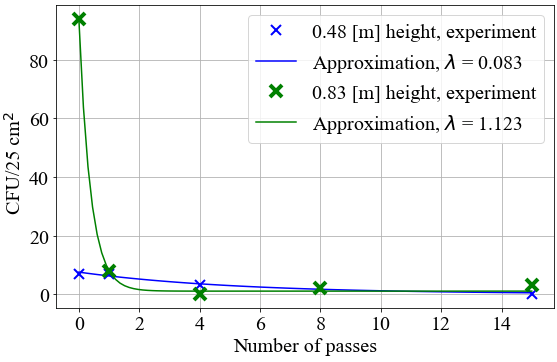}
        \caption{0.9 m distance to the disinfected shelves}
        \label{tbc_09m}
    \end{subfigure}
\caption{Results for multi-pass disinfection in grocery store.}
\vspace{-1.5em}
\label{grocery}
\end{figure}

By taking a TBC decrease rate as a performance evaluation, the efficiency is higher when the disinfected object's height (0.65 m) is close to the UV-C lamp center. The highest performance was achieved in the case of 0.35 m distance and 0.65 m height. It allows us to conclude that disinfection efficiency is higher when the object to be disinfected is close to both the robot and the center of its UV-C lamp. Finally, the optimal count of passes needed to disinfect an area should be more than 10 for 400 \textmu{W/cm$^2$} direct UV-C irradiation.

The second series of tests were proposed with turned-on DC fans in the grocery store at two heights of 0.83 m (center of a UV-C lamp), and of 0.48 m (UV-C lamp bottom level). The speed of the robot was 0.14 m/s. As a disinfection area, grocery shelves with tableware were used. Experimental results are presented in Fig. \ref{grocery}.

In general, active air circulation decreases TBC and, therefore, the performance of disinfection. However, the performance dependence in distances and heights is not the same as in the first series of tests performed without active air circulation. The highest decrease rate for 0.48 m height corresponds to 0.6 m distance from the robot to the disinfected objects. For the height of 0.83 m, the highest decrease rate is for 0.9 m distance. It is 2.04 times bigger than for the 0.3 m distance, but it gives only 20\% increase in comparison with $\lambda$ value for 0.6 m distance. Both cases (0.6 and 0.9 m distances) show approximately 1 bacteria left even after 2 robot passes (Fig. \ref{tbc_06m}, \ref{tbc_09m}). Based on this evidence, we can conclude that the most optimal distance to the target object from the robot to carry out disinfection of objects placed at different heights is 0.6 m. Moreover, the optimal number of robot passes with turned-on fans equals 2.

\subsection{Static UV-C disinfection range performance}

Several samples were taken at an industrial warehouse for the disinfection range performance analysis before and after 10 minutes of disinfection (the robot was set immobile). The swab samples for the TBC study were obtained from surfaces of cardboard boxes placed at different distances and heights. This study has been conducted in the Center of Hygiene and Epidemiology of St. Petersburg\footnote{https://78centr.ru/}, Russia. 
According to the obtained results, TBC decreased to 93\% at the distance of 1 meter from the robot. At the longer distances, an average TBC decrease was 90\%, 62\%, and 63\% of 2.8, 5.0, and 10.0 meters, respectively. There is a significant decrease in the UV-C effectiveness between ranges of 2.8 and 5.0 meters. According to this fact, we define that the UV-C range limit for the most effective disinfection (about 90\%) equals 2.8 meters in static mode.

\subsection{Experiments in dynamic condition of UV-C disinfection}
\label{range}
Experiment in dynamic condition was conducted when robot moved along the predefined trajectory. The goal of experiment was to identify UV-C disinfection range performance during robot navigation in premises.
The experiment was conducted in a grocery store, where TBC was registered at 2 grocery shelves before and after 1 robot pass. The upper and lower shelves were located at a height of 0.48 and 0.83 meters, correspondingly. The robot speed was 0.5 km/h. In the experiment, the mentioned shelves were disinfected by the robot, passing at different distances from them. Raw experimental results on registered TBC and TBC decrease mean values for upper and lower shelves are provided in Table \ref{tab:tbc_reduction}. 

\begin{table}[h!]
    \centering
    \caption{Experimental results on TBC}
    \resizebox{\columnwidth}{!}{%
    \begin{tabular}{|c|c|c|c|c|}
    \hline
    \multicolumn{1}{|l|}{Distance from} & \multicolumn{1}{c|}{Grocery} & \multicolumn{2}{c|}{Total Bacteria Count} & \multicolumn{1}{c|}{Mean TBC} \\
    \cline{3-4}
    UV-C lamps & Shelves & Before pass & After pass & decrease, \%\\
    \hline
    \multirow{2}{*}{0.3 [m]} & Upper & 18 & 0 & \multirow{2}{*}{100.00} \\
    \cline{2-4}
    & Lower & 15 & 0 & \\
    \hline
    \multirow{2}{*}{0.6 [m]} & Upper & 12 & 0 & \multirow{2}{*}{86.36} \\
    \cline{2-4}
    & Lower & 11 & 3 & \\
    \hline
    \multirow{2}{*}{0.9 [m]} & Upper & 94 & 2 & \multirow{2}{*}{84.65} \\
    \cline{2-4}
    & Lower & 7 & 2 & \\
    \hline
    \end{tabular}
    }
    \label{tab:tbc_reduction}
\end{table}

According to the results, no bacteria were left at 0.3 m distance from UV-C lamps after a single robot pass. In order to evaluate the disinfection robot performance, the following methodology was applied. First, the raw experimental results were normalized for each shelf placed on a corresponding distance from UV-C lamps. After that, the normalized data from both shelves was summarized per each distance. Finally, TBC decrease was measured in percentage and computed according to the following equation:

\begin{equation}
   TBC_{decrease} = \frac{TBC_{before}  - TBC_{after}}{TBC_{before}} \cdot 100\%.
\end{equation}

\begin{figure}[ht]
\centering
\includegraphics[width=0.49\textwidth]{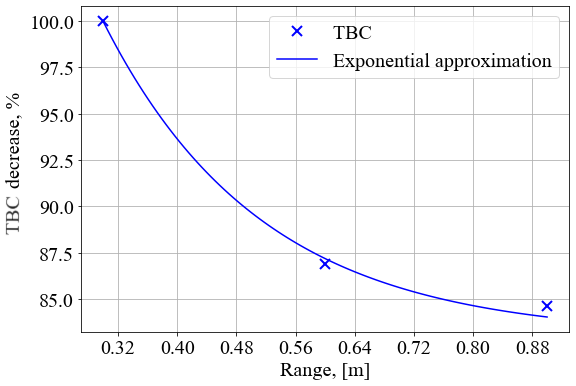}
\caption{Experimental results in dynamic conditions.}
\label{fig:dynamic_tbc_decrease}
\end{figure}

TBC decrease values are provided in Table \ref{tab:tbc_reduction} and Fig. \ref{fig:dynamic_tbc_decrease}. According to the results, it can be concluded that the robot provides up to 84\% performance in killing bacteria on surfaces located at distances less than 0.9 m from the UV-C lamps.

\vspace{-1em}
\subsection{Determination of the robot effective disinfection zone}

\begin{figure}[ht]
\vspace{-1em}
\centering
\includegraphics[width=0.45\textwidth]{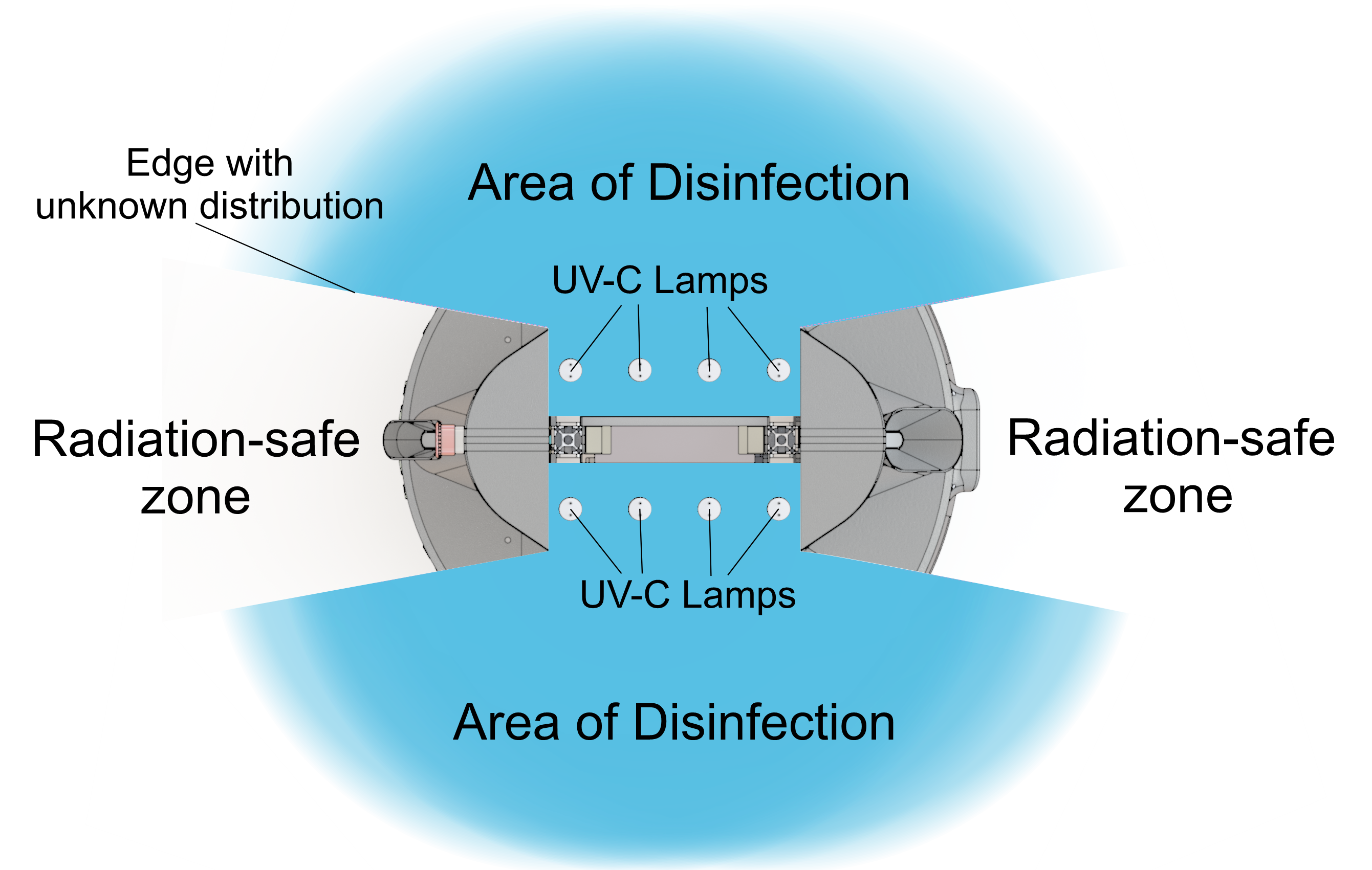}
\caption{UV-C light distribution around the robot.}
\label{fig:area_light}
\vspace{-0.5em}
\end{figure}

The proposed disinfection robot has a special, non-standard UV-C lamp arrangement. Therefore, there are zones where the distribution of disinfectant light is non-trivial (Fig. \ref{fig:area_light}). A robot disinfection zone must be identified to produce optimal path planning and provide full disinfection in selected areas. It is important to achieve both disinfection effectiveness and safety for human while irradiating strong UV-C light.

\begin{figure}[ht]
\centering
\includegraphics[width=0.45\textwidth]{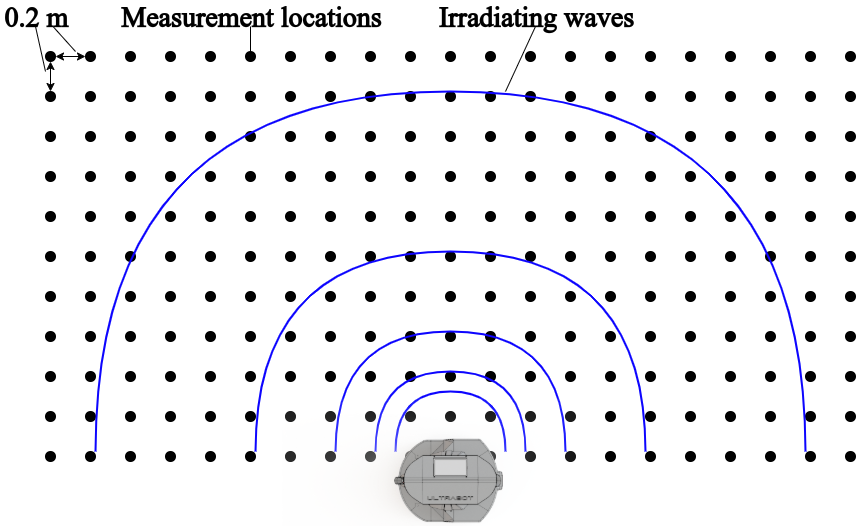}
\caption{Experimental area (Top view).}
\vspace{-1.5em}
\label{fig:experimental_area_light}

\end{figure}

In order to determine the robot's effective disinfection zone, we installed 4 fluorescent lamps with the same specifications into the robot in the original positions as UV-C ones to investigate luminosity distribution depending on the location relative to the lamps. To register luminosity values in each location, we used 3 smartphones with light sensors and installed the luminosity photometer program ``Lux Meter''.
The measurements were taken at locations on the one robot side, distributed with 0.2 m step, due to the symmetric structure of the robot (see Fig. \ref{fig:experimental_area_light}).

\begin{figure}[ht]
\vspace{-0.5em}
\centering
\includegraphics[width=0.45\textwidth]{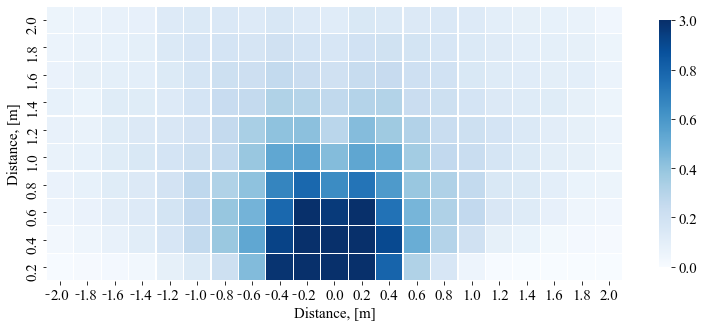}
\vspace{-0.5em}
\caption{Luminosity heatmap: the sum of normalized data obtained from 3 smartphones.}
\vspace{-0.5em}
\label{fig:luminocity_heatmap}
\end{figure}

The experimental results, which are a sum of normalized luminosity data obtained by each of the smartphones, are represented as a heatmap in Fig. \ref{fig:luminocity_heatmap}. 
According to the heatmap, light distribution is not uniform and depends on interference between four lined-up lamps, reflective sheet behind them and actual window which limits the light flow. This result confirms that the mechanical design of the robot significantly affects the light distribution of the lamps installed in it.

\begin{figure}[ht]
\centering
\includegraphics[width=0.49\textwidth]{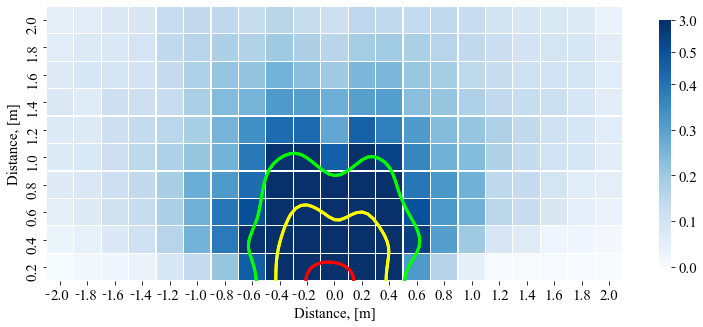}
\vspace{-1.0em}
\caption{Effective disinfection zones with outer borders - 100\% (red), 86.9\% (yellow) and 84.6\% (green).}
\vspace{-1.5em}
\label{fig:luminocity_heatmap_plus_contour_plus_zones}
\end{figure}

A luminosity threshold value should be determined to restore the exact disinfection zone and to create an outer contour of the effective UV-C irradiation area. According to the experimental results on the disinfection range, provided in Fig. \ref{fig:dynamic_tbc_decrease} and Table I, the TBC decrease in front of the UV-C lamps was 84.6\% at 0.9 m distance, which is a high value for distant disinfection according to Philips data sheet \cite{Philips}. Thus, a normalized and then summarized luminosity value equals 0.54 that corresponds to 0.9 m distance from the robot fluorescent lamps is taken as a threshold value to reconstruct the outer border contour of the effective disinfection zone, provided in Fig. \ref{fig:luminocity_heatmap_plus_contour_plus_zones} by the green line. 

Based on Fig. \ref{fig:dynamic_tbc_decrease}, TBC decrease is a volatile value dependent on the distance from the side UV-C lamps. Thus, the entire robot working area can be computed.
According to the lamp efficiency level, information on the distribution of disinfection zones will allow the robot to manage the disinfection process most effectively and achieve the desired disinfection level of the target premises. For this purpose, we also assigned the threshold luminosity values for different disinfection performances from Table I and plotted them with the red line for 100\% and the yellow line for 86.9\% effectiveness. The effective disinfection zone distribution, with the ranges obtained in \autoref{range}, is provided in Fig.  \Ref{fig:luminocity_heatmap_plus_contour_plus_zones}.

The closer zone boundaries to the center of lamps the more circular they are. It relates to the radiance from all lamps is equally high at every point of the closest disinfection zone. On the other hand, the further away from the robot, the stronger the dip in the center of the disinfection zone shape appears. This effect can be caused by an interference, in which two waves superpose to form a resultant wave of greater, lower, or the same amplitude.

\begin{figure}[ht]
\vspace{-0.5em}
\centering
\includegraphics[width=0.49\textwidth]{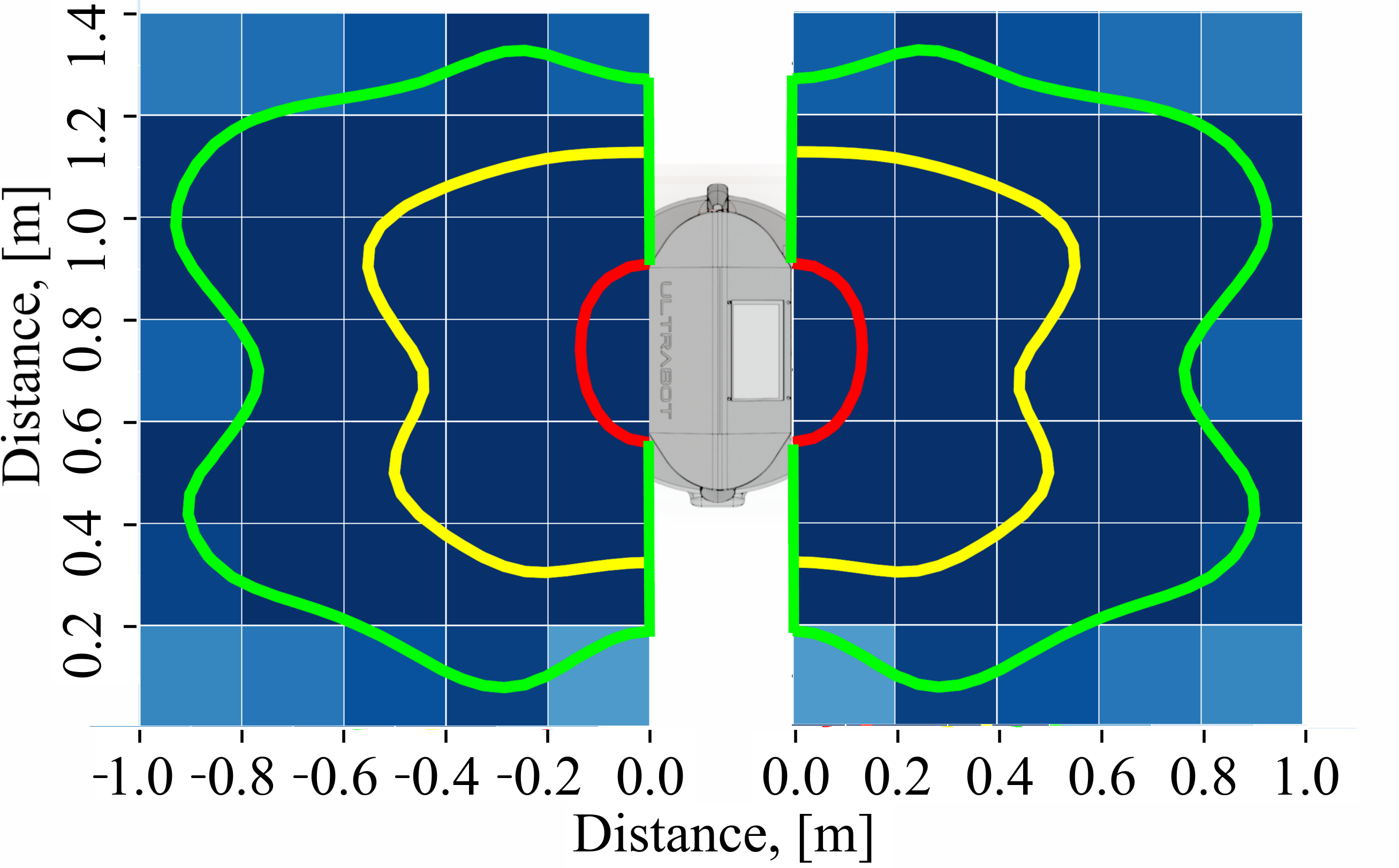}
\vspace{-0.5em}
\caption{Final shape of the robot disinfection zone.}
\vspace{-0.5em}
\label{fig:bf}
\end{figure}

Since UltraBot has a symmetrical design, the shape of the disinfection zone obtained from one side is also valid for the other side. Therefore, the final shape of a disinfection area resembles  butterfly (see Fig. \ref{fig:bf}).

\section{Conclusions}
This paper presents an autonomous UV-C disinfection robot that can work in populated indoor places. 
Robot features two-sided UV-C lamps, air ventilation, differential drive platform, and SLAM technology for autonomous navigation. 

To validate the disinfection performance of our autonomous robot, we conducted a series of qualitative experiments in a grocery store, warehouse, and souvenir shop. Experiments have proven the effectiveness of UltraBot for disinfection. The experimental results revealed actual performance for the multi-pass technique that provides 1-log reduction with combined direct UV-C exposure and ozone-based air purification after two robot passes at a speed of 0.14 m/s. This technique has the same performance as ten minutes static disinfection within 0.4 meters from UV-C lamps. 

In addition, due to the complex structure of our robot, we experimentally found the non-trivial shape of the effective disinfection zone, which turned out to be in the shape of a butterfly. It should be noted that finding the shape of the disinfection zone for autonomous robot is one of the priority task, since it must be taken into account for robot path planning.

\section{Future Work}

Our further research will be devoted to developing a genetic-based algorithm for path planning of UltraBot in an unknown dynamic environment. One of the main tasks to be solved is to predict human behavior for operative replanning of the robot trajectory without exposing people to UV-C irradiation. Therefore, the algorithm for human detection will be implemented. Additionally, the heat map of human activity will be constructed. It will allow representing the areas with higher risk of virus presence on the surfaces and in the midair. Therefore, the optimal robot path planning for more effective disinfection of such areas can be achieved. 

\bibliographystyle{IEEEtran}
\bibliography{ppcr-refs}

\end{document}